\documentclass[fleqn,twoside]{article}
\usepackage[headings]{espcrc2}

\readRCS
$Id: espcrc2.tex,v 1.2 2004/02/24 11:22:11 spepping Exp $
\usepackage{graphicx, balance}
\usepackage{amsmath}
\usepackage{float}

\usepackage[figuresright]{rotating}

\title{\textbf{Region and Location Based Indexing and Retrieval of MR-T2 Brain Tumor Images}}

\author{Krishna A N\address[DCSE]{Associate Professor, Department of Computer Science and Engineering, S J B Institute of\\ ~Technology, Bangalore-560 060, India. Contact:krishna12742004@yahoo.co.in \\},
Dr. B G Prasad\address{Professor and Head, Department of Computer Science and Engineering, B N M Institute of \\ ~Technology, Bangalore-560 070, India.
Contact: drbgprasad@gmail.com}}

\runtitle{Region and Location Based Indexing and Retrieval of MR-T2 Brain Tumor Images}
\runauthor{Krishna A N and Dr. B G Prasad}

\begin{document}
\begin{abstract}

In this paper, region based and location based retrieval systems have been implemented for retrieval of MR-T2 axial 2-D brain  images. This is done by extracting and characterizing the tumor portion of 2-D brain slices by use of a suitable  threshold computed over the entire image. Indexing and retrieval is then performed by computing texture features based on gray-tone spatial-dependence matrix of segmented regions. A Hash structure is used to index all images. A combined index is adopted to point to all similar images in terms of the texture features. At query time, only those images that are in the same hash bucket as those of the queried image are compared for similarity, thus reducing the search space and time. \\\\
{\bf Keywords :} Content Based Retrieval, Indexing, Segmentation, Texture.
\end{abstract}

\maketitle

\section{INTRODUCTION}
In medical field, a large number of diverse radiological and pathological images in digital format are generated everyday in hospitals and medical centers with sophisticated image acquisition devices and digital scanners. Medical images are generally complex in nature and are used for diagnosis, therapy, research and education. Support of prior image references is critical to radiologists or physicians current examination of images. To support their prior image reference needs, the generated images need to be processed and organized so that efficient retrieval of similar images for a current examination image is achieved. 
\vskip 2mm
Content-Based Image Retrieval (CBIR) has been initially proposed to overcome the problem caused by the subjectivity of a users perception in Text-Based Image Retrieval (TBIR). CBIR is more challenging in medical domain due to the complex nature of images. In medical domain, visual features between normal and pathological images may have only subtle differences; these may not be captured by traditional feature extraction such as color, texture or shape based on entire images. The main reason is that, important features in biomedical images are often local features of pathological regions or lesions, rather than global features of entire image. Generating local features is much more complex than global features; however, it can describe fine details of the images and allow efficient retrieval of relevant images based on local object properties. To extract regional or local features, segmentation is very important in medical imaging and generally treated as a pre-processing step. 
\vskip 2mm
Manual segmentation is a very time-consuming task and not feasible in real-time needs. Moreover, results from manual operations are not repeatable and suffer from intra-observer and inter-observer variability. In the past few decades, researchers have proposed many effective algorithms to perform automated segmentation. The successful implementation of modern mathematical and physical techniques, such as Bayesian’s analysis, template matching and deformable models, greatly enhances the accuracy of segmentation results. Compared with common image segmentation algorithms, the ones used for medical images need more concrete backgrounds and must satisfy the complex practical requirements. Due to diverse reasons, medical images are usually noisy and blurred. Effective algorithms must be robust to extract the correct information. A comprehensive survey of the current segmentation algorithms for medical images can be found in \cite{37}\cite{38}. Reliable segmentation of brain tumors is of great importance for surgical planning and therapy. Diligent efforts have been made for tumor segmentation including classification and active contour methods. 
\vskip 2mm
Supervised classification methods \cite{Dickson}\cite{David}\cite{Lao}\cite{Moon} perform segmentation using classifiers built on training data with the assumption that the statistical information extracted from the training samples can cover the testing data. Though these methods have achieved promising results, they may suffer from the inconsistency between training and testing samples due to noise and anatomy difference. Unsupervised classification methods \cite{Fletcher}\cite{Kaus}\cite{Liu}\cite{Nie}\cite{Ahmed} perform the segmentation using its specific intensity information. In order to compensate for the lack of training data, these methods iterate between segmenting the image and characterizing the properties of the each class. These methods typically need appropriate and reliable initialization of number of clusters to get good results. 
\vskip 2mm
Active contour methods perform segmentation utilizing both intensity and geometrical information of objects to be segmented \cite{Ho}\cite{Cobzas}\cite{Ayed}\cite{Taheri}. These are model-based techniques for delineating region boundaries by using closed parametric curves or surfaces that deform under the influence of internal and external forces. To delineate an object boundary in an image, a closed curve or surface must first be placed near the desired boundary and then allowed to undergo an iterative relaxation process. Internal forces are computed from within the curve or surface to keep it smooth throughout the deformation. External forces are usually derived from the image to drive the curve or surface towards the desired feature of interest. Level-set methods were introduced to deformable models by casting the curve evolution problem in terms of front propagation rather than energy minimization.
\vskip 2mm
In this paper, we have proposed and implemented a novel method for segmentation of brain tumors and retrieval based on segmented regions. In the context of CBIR, after manual or automatic segmentation of an image, the resulting segments are termed as regions of interest (ROI). ROI-based retrieval methods extract features of the segmented regions and perform similarity comparisons at the granularity of the region. The main objective of using region features is to enhance the ability of capturing as well as representing the focus of the user’s perceptions of image content \cite{Jinman}\cite{W. Liu}\cite{Feng Jing}\cite{Khanh Vu}. Since medical images are often highly textured and the one based on gray-tone spatial-dependence matrix is proved to be powerful in texture analysis, so it is adopted to describe the texture feature of the segmented region.

\section{REGION-OF-INTEREST SEGMENTATION}
Image segmentation by use of a suitable threshold is one of the main techniques. It is the oldest and still most commonly used technique because of its simplicity and efficiency. Many methods \cite{Otsu}\cite{Kanpur}\cite{Sahoo}\cite{Pal}\cite{Cheriet}\cite{Debashis} have been proposed in the literature to find one or more thresholds from the histogram of an image and perform the segmentation based on  the threshold. The proposed segmentation algorithm uses the global threshold selection method which uses gray-level distribution. A global threshold is the that partitions the entire image with a single threshold value. After the thresholding, region labeling algorithm is applied to obtain clusters of different sizes. The cluster of largest size is considered as tumor (region of interest). An image can be represented by a 2-D gray-level intensity function $f(x, y)$. The value of $f(x, y)$ is the gray-level ranging from 0 to $L-1$, where $L$ is the number of distinct gray-levels. The major steps of the proposed segmentation algorithm is shown in Table 1.

\begin{table*}
{\caption{Segmentation Algorithm by use of a Global Threshold}}
\begin{center}
\begin{tabular}{|l|} \hline
1. Compute $T$ by iterative threshold selection algorithm (shown in Table 2) \\
2. Find $t^*$, the maximum between-class variance  \\
3. The optimal threshold $T^*$ is defined by the sum $T$ and $t^*$ \\ 
4. The segmented image, $g(x, y)$ is given by \\
	$\hspace{1.1 cm} g(x, y) = \left\{ 
\begin{array}{l l}
  1 & \quad \mbox{if $f(x, y) > T^*$}\\
  0 & \quad \mbox{if $f(x, y) \le T^*$} \hspace{3.0 cm} \\
\end{array} \right.$ \\
\hspace{0.5 cm} where $T^*$ is a constant applicable over an entire image \\
5. Labelling of segmented image is done for eliminating small clusters \\ \hline
\end{tabular} 
\end{center}
\end{table*} 

\begin{table*}
{\caption{Algorithm: To Find $T$ Iteratively}}
\begin{center}
\begin{tabular}{|l|} \hline
1. Initialize $T$ = Average gray level of the image.	\\
2. Compute $\mu_1$ for pixels less than or equal to $T$. \\
3. Compute $\mu_2$ for pixels greater than $T$.	\\
4. Compute a new threshold $T = \frac{1}{2}(\mu_1 + \mu_2)$ \\
5. Repeat step 2 through 4 until the difference \\
\hspace{0.3 cm} in $T$ in successive iterations is smaller \\ 
\hspace{0.3 cm} than a predefined $T_0$ (= 0).	\\ \hline
\end{tabular} 
\end{center}
\end{table*} 

\subsection{Algorithm: To Find $t^*$ by Otsu Method}
Suppose that there are $N$ pixels and $L$ gray levels ($0$, $1$, ..., $L-1$) in an image. Let $n_l$ denote the number of pixels at level $l$, then $N = \sum_{l=0}^{l-1} {n_l}$. The histogram of an image can be normalized as a probability distribution by 
\[\hspace{1 cm} P_l = \frac{n_l}{N}, \hspace{1 cm} \sum_{l=0}^{l-1} {P_l} = 1 \]
Assume that a threshold $t$ divides the gray levels into two clusters $S_1$ = {0. 1, ..., $t$} and $S_2$ = {$t$ + 1, $t$ + 2,..., $L - 1$} . Let $\sigma_{B}^{2}(t)$ be the between-class variance of the gray levels \cite{Otsu}. Then the optimal threshold $t^*$ is obtained by
\[ t^* = \underset{0 \le t < L-1}{\operatorname{arg\,max}} \hspace{0.1 cm} {\sigma_B^2{(t)}} \]
where
\[ \sigma_B^2{(t)} = \sum_{l=1}^{2}{w_i(\mu_i - \mu_T)^2}, \hspace{1.0 cm} w_1 = \sum_{l=0}^{t}{P_l}\]
\begin{minipage}{0.05in}
\end{minipage}
\[ w_2 = \sum_{l=t+1}^{L-1}{P_l}, \hspace{1.0 cm} \mu_1 = \sum_{l=0}^{t}{\frac{l P_l}{w_1}}\]
\begin{minipage}{0.05in}
\end{minipage}
\[ \mu_2 = \sum_{l=t+1}^{L-1}{\frac{l P_l}{w_2}}, \hspace{1.0 cm} \mu_T = \sum_{l=0}^{L-1}{l P_l} \]

\subsection{Labeling of the Segmented Image}
The aim is to find each connected region of pixels that were detected by use of a threshold and give all the pixels in that region their own unique label. Also count the number of pixels in each region. The simplest and the most common labeling algorithm scans the image pixel by pixel, invoking a recursive labeling procedure whenever a non-zero pixel is found. This is done by starting at the pixel and propagating to any of the 8-neighbors that were also detected by use of a threshold. The pixels visited in the input image has its value set to zero so that it cannot be visited again by the labeling procedure. At the end of the procedure, all the pixels belonging to the region have been set to 0 in the input image, making them indistinguishable from the background and the corresponding pixels in the output image have been assigned a region number. The region number is then incremented, ready for the next connected region. The recursive labeling procedure is given in Table 3. Table 4 shows representative snapshots depicting tumor identification of MR-T2 brain images.
\begin{table}
{\caption{Algorithm: Labeling Segmented Image}}
\begin{center}
\begin{tabular}{|l|} \hline
int n = 1;	\\
for (int x = 0; x $<$ iw; ++x)	\\
\hspace{0.5 cm}	for (int y = 0; y $<$ ih; ++y)	\\
\hspace{1 cm} if (in.getSample(x, y, 0) $>$ 0) \{    \\      			
\hspace{1.5 cm} label(in, out, volume, x, y, n);   \\
\hspace{1.5 cm} ++n;		        \\
\hspace{1 cm} \}       	     \\ 

label(in, out, volume, x, y, n) \{	\\
\hspace{0.5 cm} in.setSample(x, y, 0, 0);	\\
\hspace{0.5 cm} region[x][y] = n;	\\
\hspace{0.5 cm} volume[n]++;	\\
\hspace{0.5 cm} int j, k;	\\ 	
\hspace{0.5 cm} for(int i = 0; i $<$ 8; ++i) \{	\\
\hspace{1.0 cm} j = x + delta[i][0];	\\	
\hspace{1.0 cm} k = y + delta[i][1];   	\\
\hspace{1.0 cm} if(inImage(j,k)\&\&\\
\hspace{1.0 cm} in.getSample(j,k,0)$>$0) 	\\
\hspace{1.3 cm} label(in, region, volume, j, k, n);	\\
\hspace{1.0 cm}    \}    \\
  \} 	\\ 
  
  private final boolean inImage(int x, int y) \{		\\
\hspace{0.5 cm} return x$\ge$0 \&\& x$<$256 \&\& y$\ge$0 \&\&\\
\hspace{0.5 cm} y$<$256;	\\
  \}		\\ \hline
\end{tabular} 
\end{center}
\end{table} 

\begin{table*}
{\caption{Some Results of Tumor Segmentation}}
\centering
\begin{tabular}{|p{2 cm}|p{3 cm}|p{3 cm}|p{3 cm}|}
\hline
Query \hspace{0.5cm} Images & \includegraphics[width=0.2\textwidth]{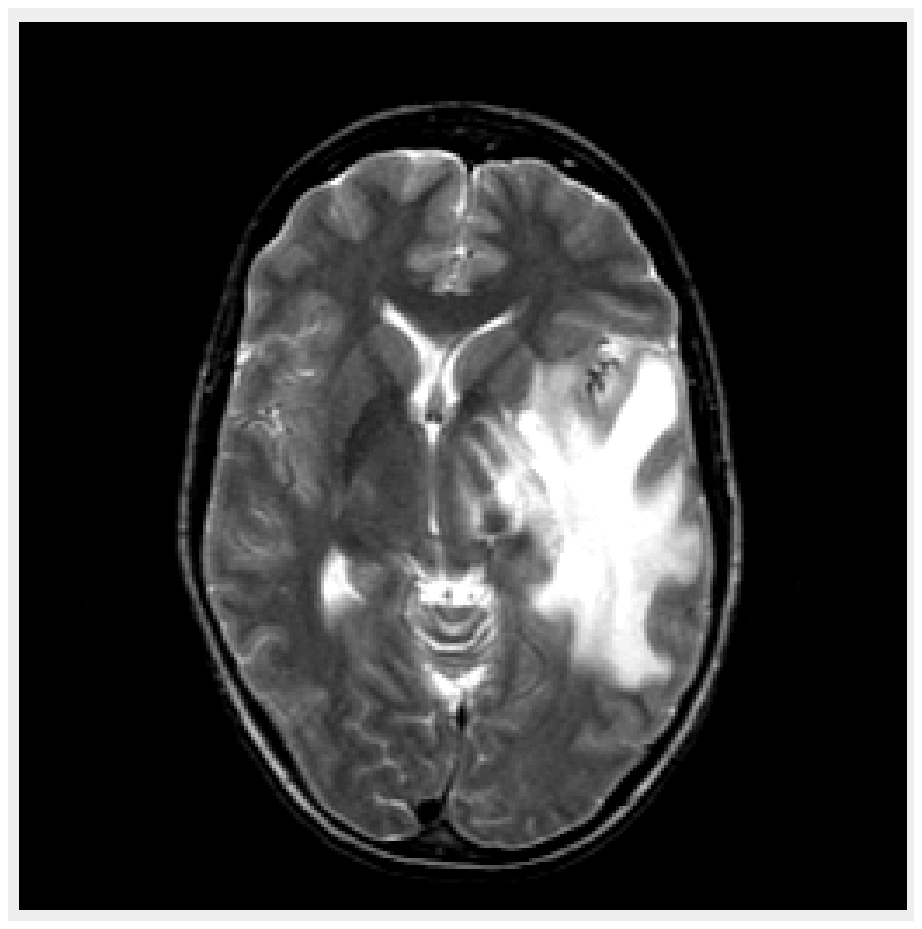} & \includegraphics[width=0.2\textwidth]{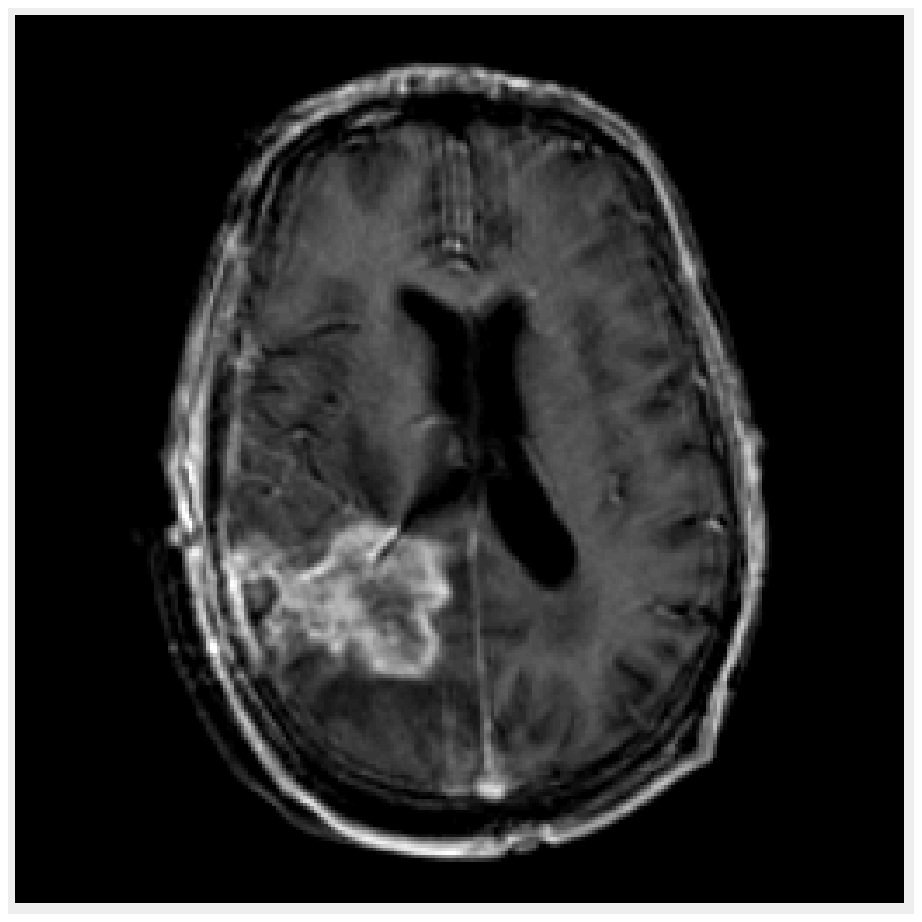} & \includegraphics[width=0.2\textwidth]{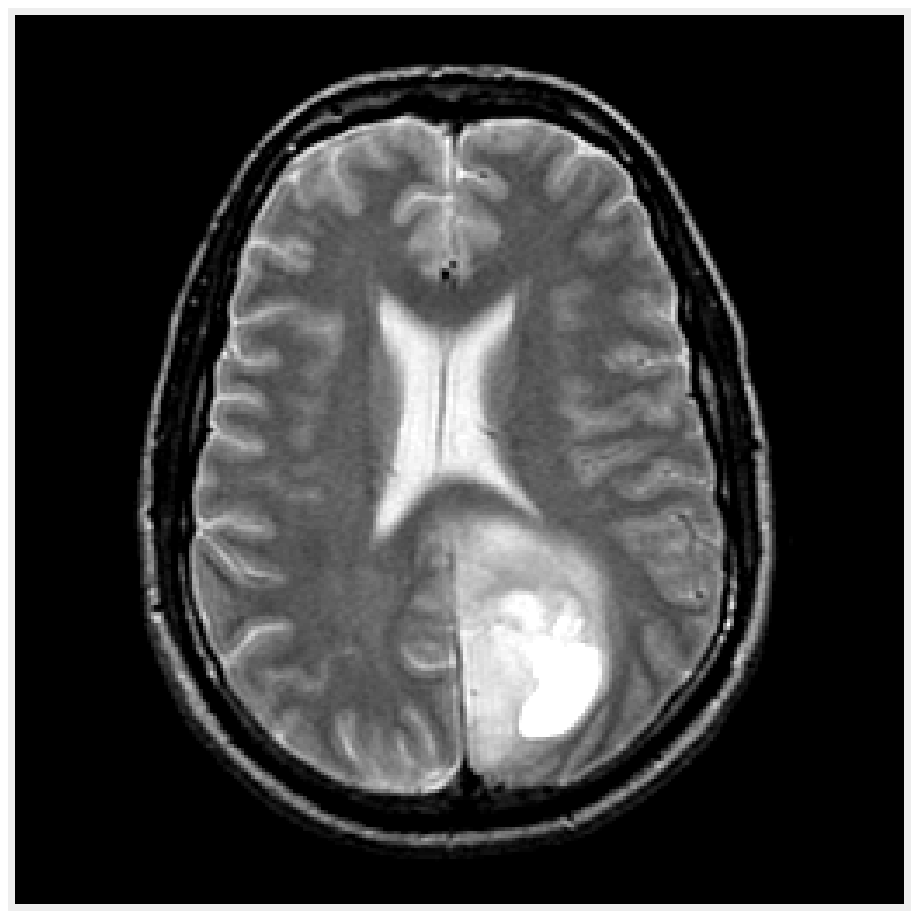} \\ \hline
Segmented Images & \includegraphics[width=0.2\textwidth]{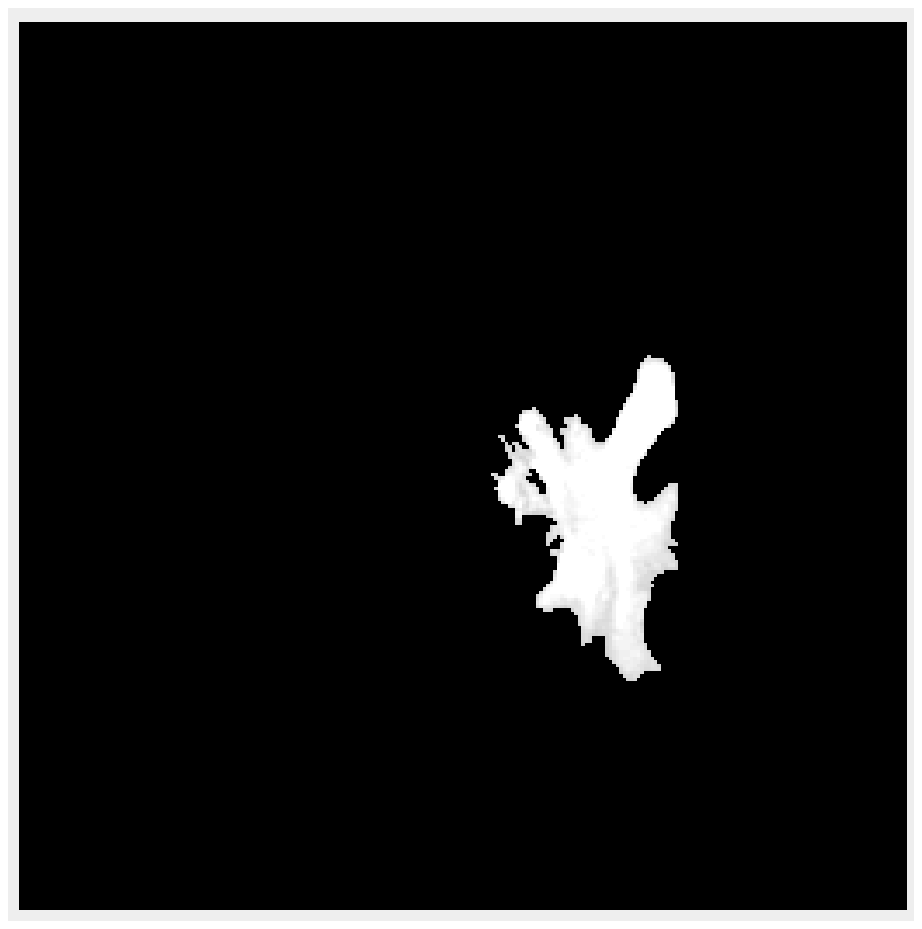} & \includegraphics[width=0.2\textwidth]{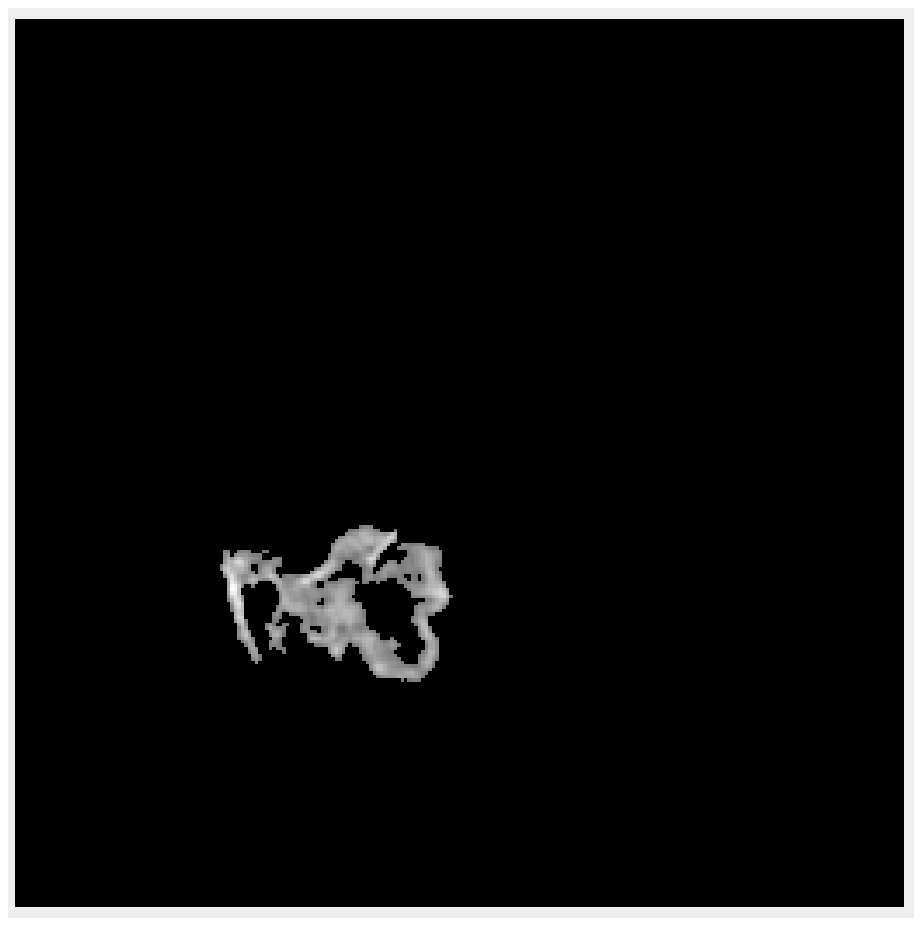} & \includegraphics[width=0.2\textwidth]{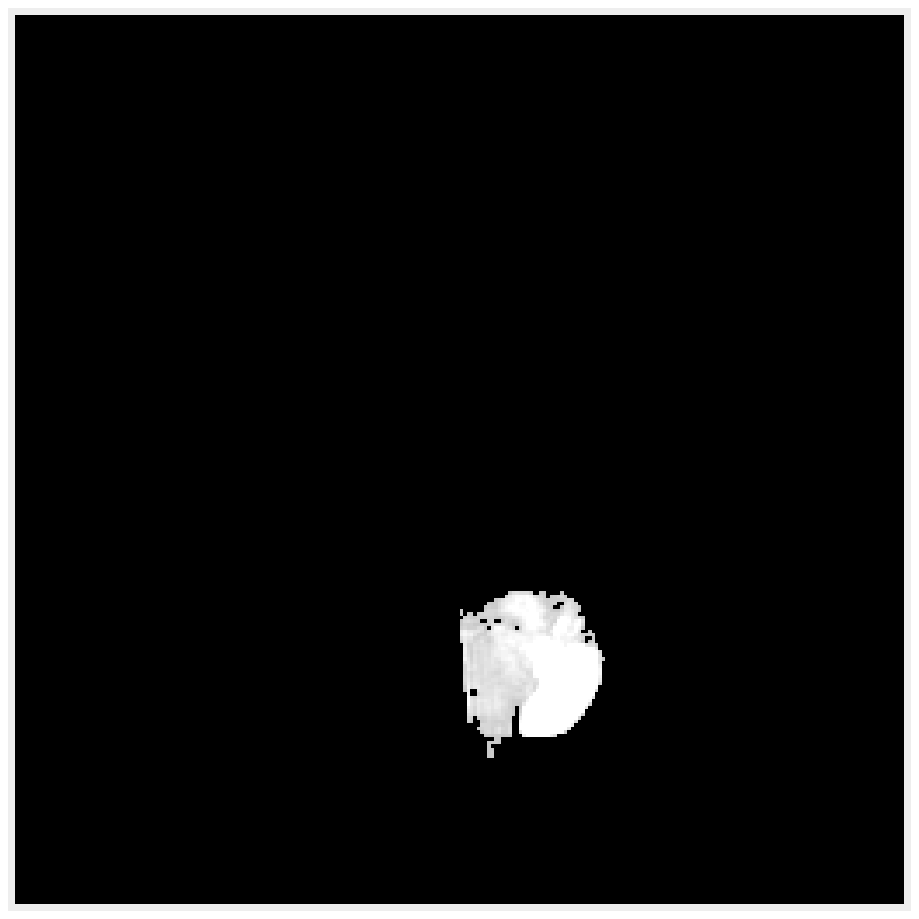} \\ \hline
\end{tabular}
\end{table*}

\section{FEATURE REPRESENTATION}
Texture information is specified by a set of gray-tone spatial-dependence matrices that are computed for various angular relationships and distances between neighboring resolution cell pairs on the image. All the textural features are derived from these angular nearest-neighbor gray-tone spatial-dependence matrices. Suppose an image to be analyzed is rectangular and has $N_x$ resolution cells in the horizontal direction and $N_y$, resolution cells in the vertical direction. Suppose that the gray tone appearing in each resolution cell is quantized to $N_g$ levels. Let $L_x$ = {1, 2, ... ,$N_x$} be the horizontal spatial domain, $L_y$ = {1, 2, ..., $N_y$} be the vertical spatial domain, and G = {1, 2, ..., $N_g$} be the set of $N_g$ quantized gray tones. The set $L_y$ x $L_x$ is the set of resolution cells of the image ordered by their row-column designations. The image $I$ can be represented as a function which assigns some gray tone in $G$ to each resolution cell in $L_y$X$L_x$; $I$: $L_y$X$L_x$ $\rightarrow$ $G$.
\vskip 2mm
Four closely related measures from which the texture features we have used are derived using angular nearest-neighbor gray-tone spatial-dependence matrices: $P(i,j,d,0^o)$, $P(i,j,d,45^o)$, $P(i,j,d,90^o)$ and $P(i,j,d,135^o)$. We assume that the texture-context information in an image $I$ is contained in the "overall" or "average" spatial relationship which the gray tones in image $I$ have to one another. More specifically, this texture-context information has been adequately specified by a matrix of relative frequencies $P_{ij}$ with which two neighboring resolution cells, one with gray tone $i$ and the other with gray tone $j$ separated by distance $d$ occur on the image. Such matrices of gray-tone spatial-dependence frequencies are a function of the angular relationship between the neighboring resolution cells as well as a function of the distance between them. Formally, for angles quantized to $45^o$ intervals, the unnormalized frequencies are defined by Eqs. (1)-(4). The details of computing these texture measures is presented in \cite{29}.
\begin{multline}
P(i,j,d,0^o) = \#\{((k,l),(m,n)) \in \\  (L_y \textnormal{ X } L_x)  \textnormal{ X } (L_y \textnormal{ X } L_x) | k-m = 0,\\  |l-n| = d,  I(k,l) = i, I(m,n) = j\} 
\end{multline}
\begin{multline}
P(i,j,d,45^o) = \#\{((k,l),(m,n)) \in \\(L_y \textnormal{ X } L_x)  \textnormal{ X } (L_y \textnormal{ X } L_x) | (k-m = d,\\ l-n = -d)   or (k-m = -d, l-n = d), \\  I(k,l) = i, I(m,n) = j\}
\end{multline}
\begin{multline}
P(i,j,d,90^o) = \#\{((k,l),(m,n)) \in \\(L_y \textnormal{ X } L_x)  \textnormal{ X } (L_y \textnormal{ X } L_x) | |k-m| = d, \\|l-n| = 0,  I(k,l) = i, I(m,n) = j\}
\end{multline}
\begin{multline}
P(i,j,d,135^o) = \#\{((k,l),(m,n)) \in \\(L_y \textnormal{ X } L_x)  \textnormal{ X } (L_y \textnormal{ X } L_x) | (k-m = d, \\l-n = d)  or (k-m = -d, l-n = -d), \\ I(k,l) = i, I(m,n) = j\} 
\end{multline}
where \# denotes the number of elements in the set. \\
\textit{Note}: These matrices are symmetric; $P (i,j; d, a) = P (j, i; d, a)$. \vskip 6pt

We compute four closely related measures $P(i, j, d, \theta)$ quantized to $45^0$ intervals with $d=1$ from which all our three texture features are derived. Out of the equations which define a total set of 14 measures of textural features \cite{29}, we have used the three most distinguishing parameters to describe the texture of an image as depicted by Eqs. (5)-(7). 
\begin{equation}
Energy = \sum_{i}\sum_{j}P(i,j)^2 \hspace{2 cm}
\end{equation}
\begin{equation}
Entropy = -\sum_{i}\sum_{j}P(i,j)log(P(i,j)) \hspace{0.3 cm}
\end{equation}
\begin{equation}
Contrast = \sum_{n=0}^{N_g-1}n^2\left\{\sum_{i=1}^{N_g}\sum_{\substack{j=1 \\ |i-j|=n}}^{N_g}P(i,j)\right\}
\end{equation}

\section{REGION BASED INDEXING AND RETRIEVAL (RBIR)}
A data structure based on hashing technique is used to store all images along with the texture feature data. A combined index is adopted to point to all similar images in terms of the texture features. When a query is made based on an example image, the example image is processed for index value. Only those images that are in the same hash bucket as those of the queried image are compared for similarity. For each image in the database, segmentation procedure discussed in section 2 is applied to identify region-of-interest and describe segmented region by texture features: \textit{entropy}, \textit{energy} and \textit{contrast}. The texture features extracted are quantized to integer values between 0 to 9. The combined index of these features is: $100 * [entropy] + 10 * [energy] + [contrast]$, where [ ] represents quantization. Each combined index stores feature data along with the image object. For a query image, after finding the region-of-interest, the above mentioned texture features have to be computed, quantized and the combined index derived. Only those images that are stored at the combined index matching those of the query index, are extracted as resultant target images for a given query image. These resultant images are sorted using Euclidean distance measure in the decreasing order of similarity against the query image and displayed four images at a time using JAVA-AWT based GUI. A few representative snapshots of region-based indexing and retrieval are shown in Figure 1. Hash table offers very fast insertion and searching. Irrespective of the size of the data, insertion and searching can take close to constant time $O(1)$. Not only are they fast, hash tables are simple and easy to implement. Searching using hash tables are significantly faster than using tree, which operate in $O(log N)$ time.

\begin{figure*}
\centering
\begin{tabular}{cc}
\includegraphics[width=0.5\textwidth]{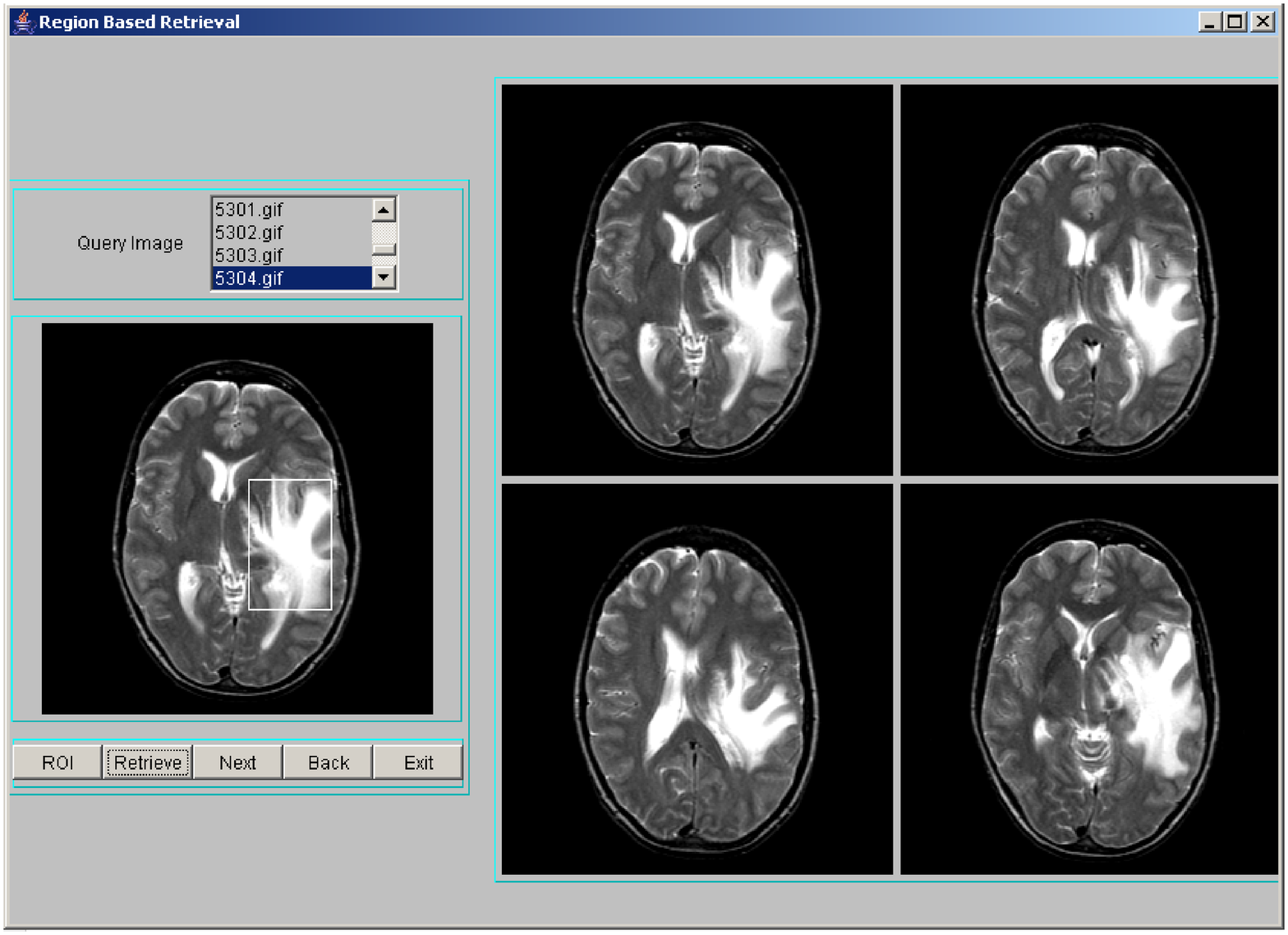} & 
\includegraphics[width=0.5\textwidth]{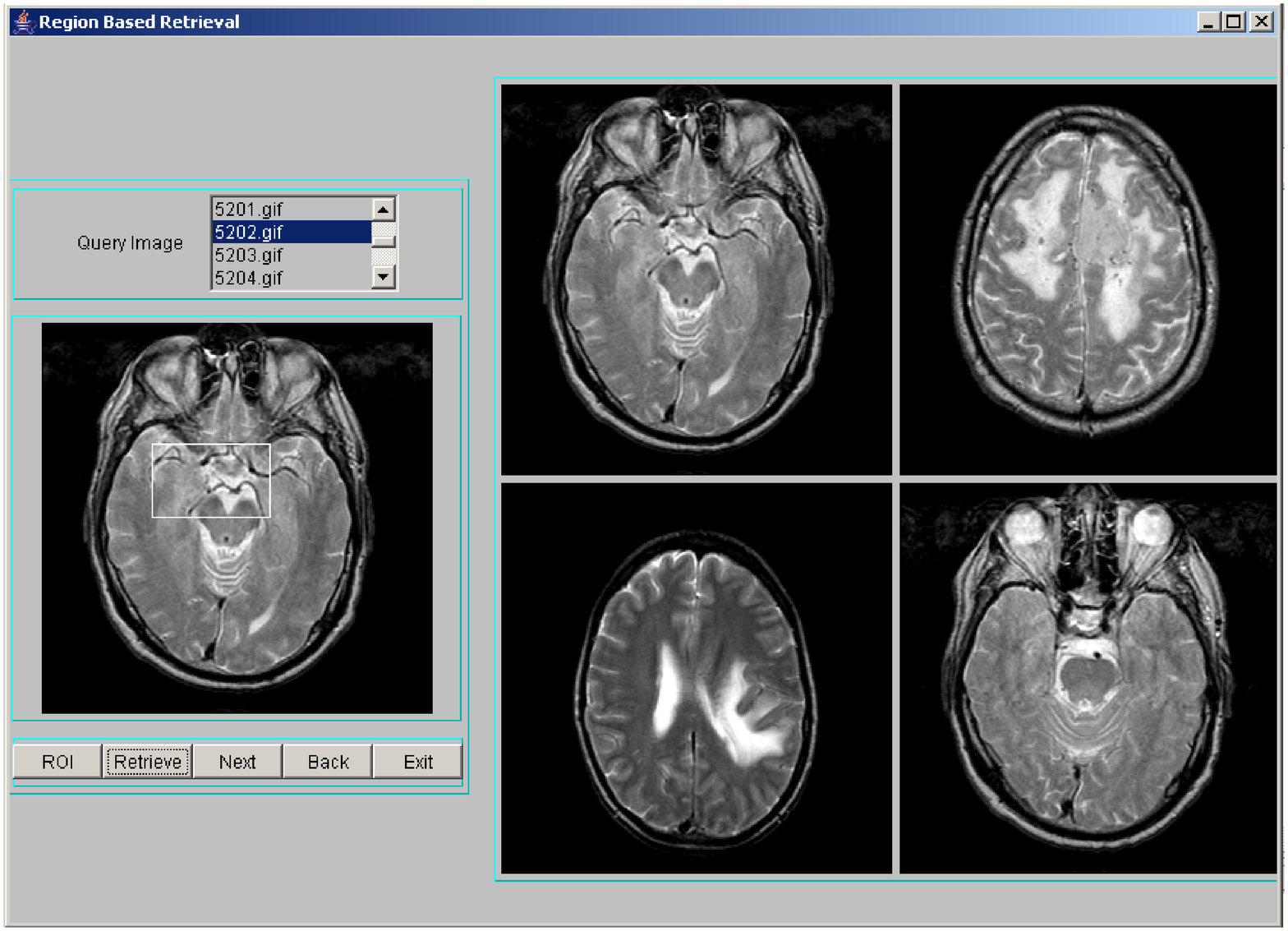} \\ 
\end{tabular}
{\caption{A Few Representative Snapshots of Region Based Indexing and Retrieval}}
\end{figure*}

\section{LOCATION BASED INDEXING AND RETRIEVAL (LBIR)}
Location Based Indexing and Retrieval is performed by finding spatial location of a segmented region. The importance of location of objects is to identify the area of involvement of tumor like sensory or motor. The brain has unique areas for speech, hearing, visual, temperature regulation etc.. If the tumor which may not be centrally located, occurs at any particular area/location, then the corresponding organ gets affected. Hence location is an important feature to be indexed. To compute the location of a region, we divide the image space into 3x3 grid cells and number them 0-8 as shown in Figure 2. The region is likely to overlap number of cells in the image space. The index assigned is the cell number that is maximally covered by the region. A program segment to find location of a region is given in Table 5. We have considered an image size of 256x256 pixels in our work. The position of a region forms the location index. For each image in the database, segmentation procedure is applied to identify region-of-interest and describe segmented region by texture features: \textit{entropy}, \textit{energy} and \textit{contrast}. Each location index stores region texture feature data along with the image object. For a query image, after finding the region-of-interest, the above mentioned texture features have to be computed and the location index is derived. Only those images that are stored at the location index matching those of the query index, are extracted as resultant target images for a given query image. These resultant images are sorted using Euclidean distance measure in the decreasing order of similarity against the query image and displayed four images at a time using JAVA-AWT based GUI. A few representative snapshots of location-based indexing and retrieval are shown in Figure 2.
\begin{figure*}
\centering
\begin{tabular}{cc}
\includegraphics[width=0.5\textwidth]{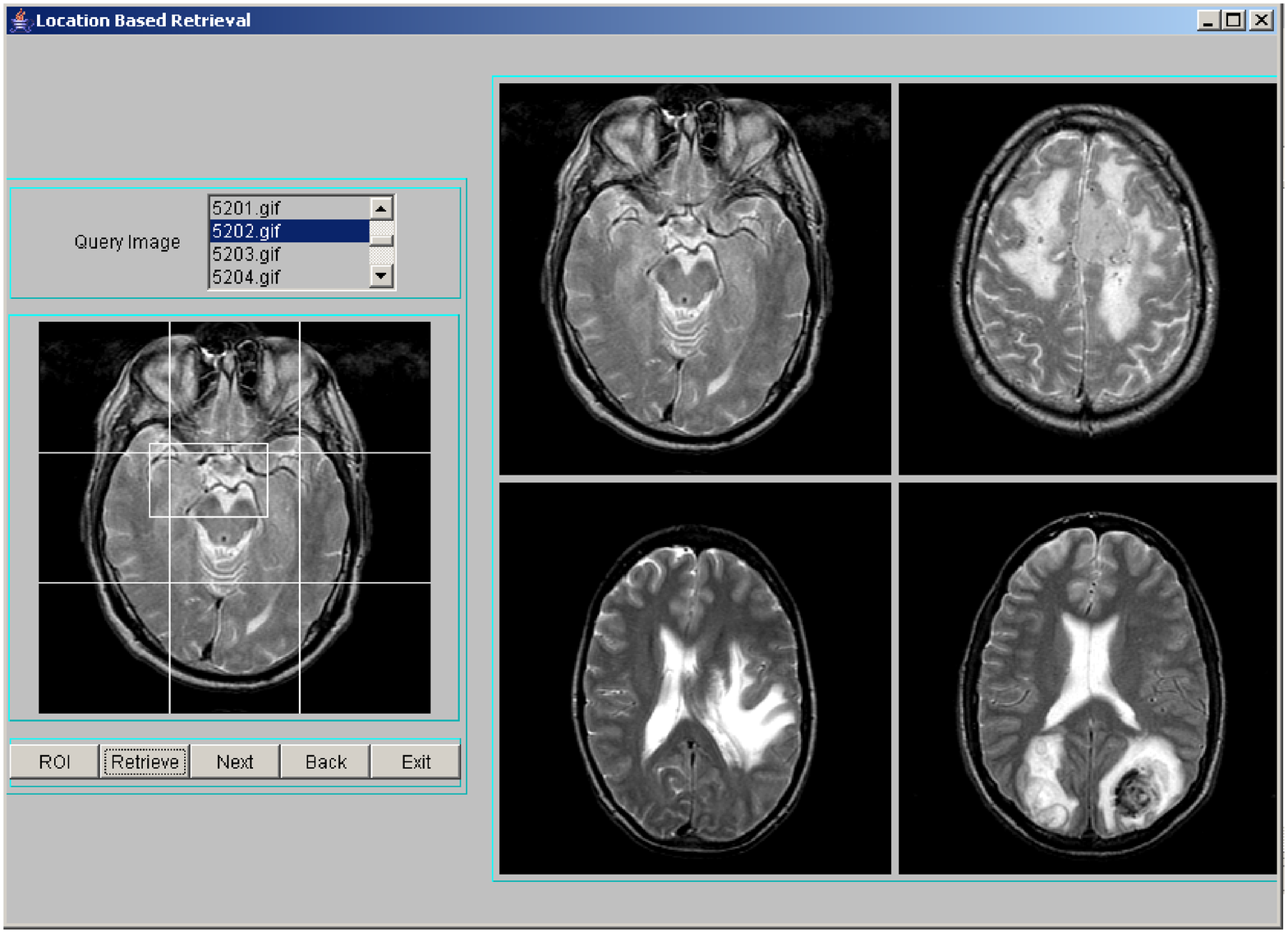} & 
\includegraphics[width=0.5\textwidth]{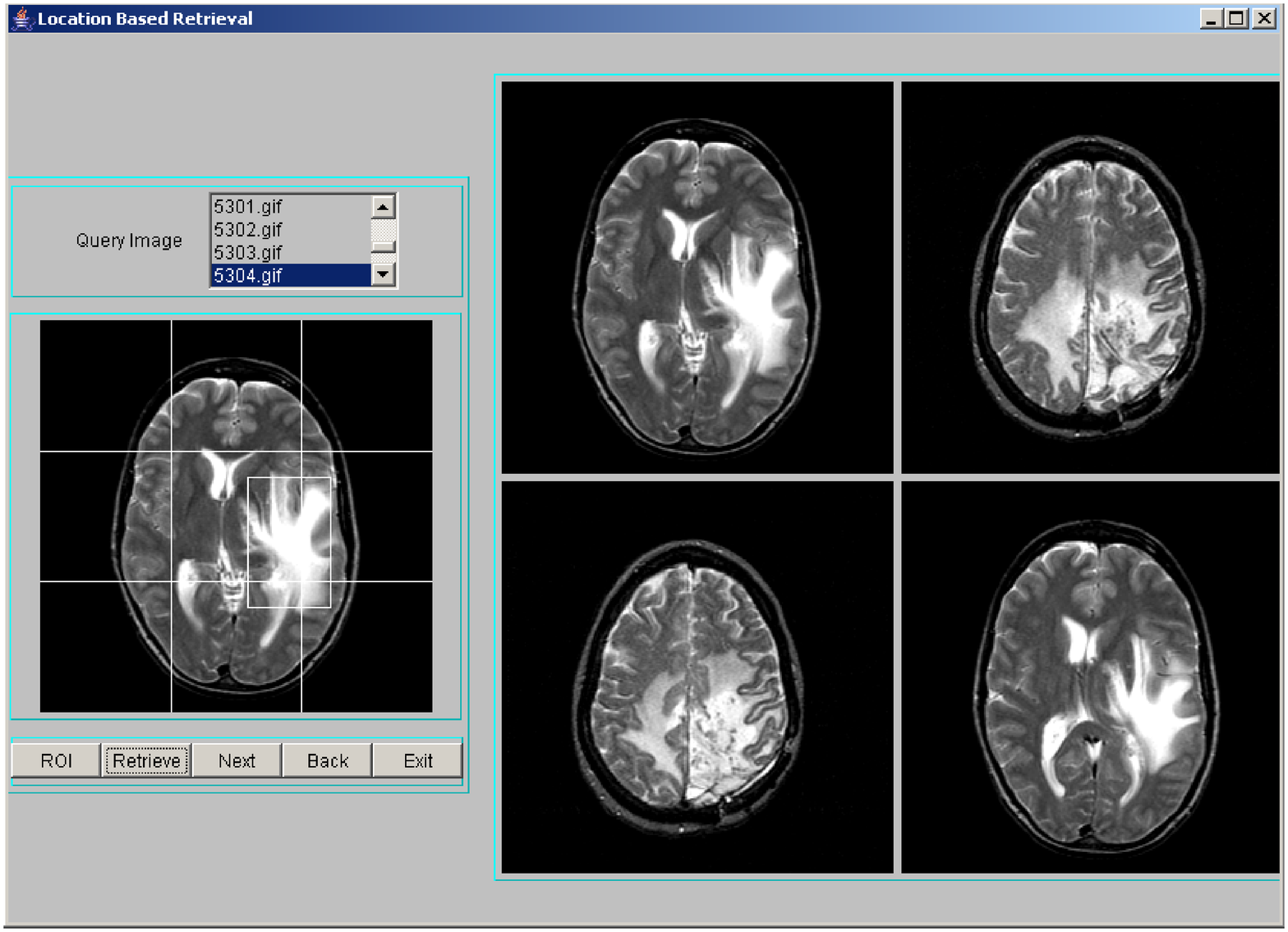} \\ 
\end{tabular}
{\caption{A Few Representative Snapshots of Location Based Indexing and Retrieval}}
\end{figure*}

\begin{table}
{\caption{Program Segment to Find the Location of a Region}}
\begin{center}
\begin{tabular}{|l|} \hline
public void findPos(x1, y1, x2, y2) \{ \\
\hspace{0.5 cm} x1, y1 and x2, y2 specifies the \\
\hspace{0.5 cm} top left corner and bottom right \\
\hspace{0.5 cm} corner position of the region \\
\hspace{0.5 cm} for which position is to be computed \\
\hspace{0.5 cm} int pos = 0, row=0, col = 0;; \\
\hspace{0.5 cm} row = findLoc(x1, x2); \\
\hspace{0.5 cm} col = findLoc(y1, y2); \\
\hspace{0.5 cm} pos = row*3 + col;	\\
\} \\ \\

int findLoc(int loc1, int loc2) \{ \\
\hspace{0.5 cm} int loc = 0;	\\
\hspace{0.5 cm} for(int i=1; i$\le$loc2/85; i++) \{	\\
\hspace{1.0 cm} if(Abs(85*i-loc2) $\ge$ Abs(85*i-loc1))	\\
\hspace{1.5 cm} loc++;   \\
\hspace{0.5 cm} \}   	\\
\hspace{0.5 cm} return loc;	\\
   \}	\\	\hline
\end{tabular} 
\end{center}
\end{table} 
	
\begin{table*}
{\caption{Retrieval Results of Percentage Precision of CBIR for Top 10 Retrievals}}
\centering
\begin{tabular}{lrrrrrrrrrr}
\hline
Classes & \multicolumn{10}{c}{Number of Retrievals} \\ \cline{2-11}
 & 1 & 2 & 3 & 4 & 5 & 6 & 7 & 8 & 9 & 10 \\ \cline{2-11}
Glioma & 100 & 100 & 100 & 75 & 80 & 66.66 & 71.42 & 75 & 66.66 & 60 \\ 
Meningioma & 100 & 100 & 100 & 75 & 60 & 66.66 & 57.14 & 50 & 44.44 & 40 \\ 
Carcinoma & 100 & 100 & 66.66 & 50 & 40 & 50 & 42.85 & 37.5 & 44.44 & 40 \\ 
Sarcoma & 100 & 50 & 66.66 & 50 & 60 & 50 & 42.85 & 50 & 44.44 & 40 \\
 & & & & & & & & & & \\ 
Average & 100 & 87.5 & 83.33 & 62.5 & 60 & 58.33 & 53.56 & 53.125 & 50 & 45 \\ \hline
\end{tabular}
\end{table*}

\begin{table*}
{\caption{Retrieval Results of Percentage Precision of LBIR for Top 10 Retrievals}}
\centering
\begin{tabular}{lrrrrrrrrrr}
\hline
Classes & \multicolumn{10}{c}{Number of Retrievals} \\ \cline{2-11}
 & 1 & 2 & 3 & 4 & 5 & 6 & 7 & 8 & 9 & 10 \\ \cline{2-11}
Glioma & 100 & 100 & 100 & 100 & 100 & 83.33 & 85.71 & 87.5 & 80 & 80 \\ 
Meningioma & 100 & 100 & 66.66 & 50 & 40 & 50 & 57.14 & 50 & 55.55 & 50 \\ 
Carcinoma & 100 & 50 & 66.66 & 50 & 60 & 50 & 57.14 & 50 & 55.55 & 50 \\ 
Sarcoma & 100 & 100 & 66.66 & 75 & 80 & 83.33 & 71.42 & 62.5 & 66.66 & 70 \\
 & & & & & & & & & & \\ 
Average & 100 & 87.5 & 75 & 68.75 & 70 & 66.66 & 67.85 & 62.5 & 66.66 & 62.5 \\ \hline
\end{tabular}
\end{table*}

\begin{table*}
{\caption{Retrieval Results of Percentage Precision of RBIR for Top 10 Retrievals}}
\centering
\begin{tabular}{lrrrrrrrrrr}
\hline
Classes & \multicolumn{10}{c}{Number of Retrievals} \\ \cline{2-11}
 & 1 & 2 & 3 & 4 & 5 & 6 & 7 & 8 & 9 & 10 \\ \cline{2-11}
Glioma & 100 & 100 & 100 & 100 & 100 & 100 & 100 & 87.5 & 77.77 & 70 \\ 
Meningioma & 100 & 100 & 66.66 & 75 & 60 & 50 & 57.14 & 50 & 44.44 & 40 \\ 
Carcinoma & 100 & 100 & 100 & 100 & 80 & 83.33 & 71.42 & 75 & 66.66 & 60 \\ 
Sarcoma & 100 & 100 & 100 & 100 & 100 & 83.33 & 71.42 & 75 & 66.66 & 60 \\
 & & & & & & & & & & \\ 
Average & 100 & 100 & 91.66 & 93.75 & 85 & 79.16 & 75 & 68.75 & 63.88 & 57.5 \\ \hline
\end{tabular}
\end{table*}

\section{PERFORMANCE ANALYSIS}
The two indexing and retrieval techniques implemented are: RBIR and LBIR. The retrieval performance is measured 
using precision and recall. We have experimented with a small database of 100 images of 4 classes, 
each of 25 images. These images are of 256 x 256 GIF format obtained from brain atlas of Harvard University.
Each of the 100 images were used as query image and performance evaluated.
Precision results are computed from the number of similar images (i.e., images belonging to the same class)
in the top 10 retrieved images. Table 6, 7 and 8 shows results depicting the \textit{Precision rates}
for 4 different classes tabulated for retrieval from top 1 to top 10 retrieved images. 
The results for RBIR is shown to be better when compared to CBIR and LBIR, which leads to almost 10 percent 
increase in precision rates. Table 9 depicts the \textit{Recall rates} for the same 4 different classes in 
the database. Here also, each of the 100 images were used as a query image and the number of matches in the 
top 20 retrieved images was counted and is shown to drastically increase recall rates by almost 10 percent. 
The precision recall graph for plotting the average precision retrieval rates for top 10 retrievals of the 
three indexing schemes is shown in Figure 3.  

\begin{table}
{\caption{Retrieval Results of Percentage Recall Rate for Top 20 Retrievals}}
\centering
\begin{tabular}{lrrr}
\hline
Classes & \hspace{0.2 cm} CBIR & \hspace{0.2 cm} LBIR & \hspace{0.2 cm} RBIR \\ \hline  
Glioma & 60 & 60 & 65   \\ 
Meningioma & 45 & 45 & 50   \\ 
Carcinoma & 45 & 45 & 50  \\ Sarcoma & 50 & 55 & 60  \\
 &  &  & \\ 
Average & 50 & 51.25 & 58.75  \\ \hline
\end{tabular}
\end{table}

\begin{figure*}
  \centering
    \includegraphics[width=0.75\textwidth]{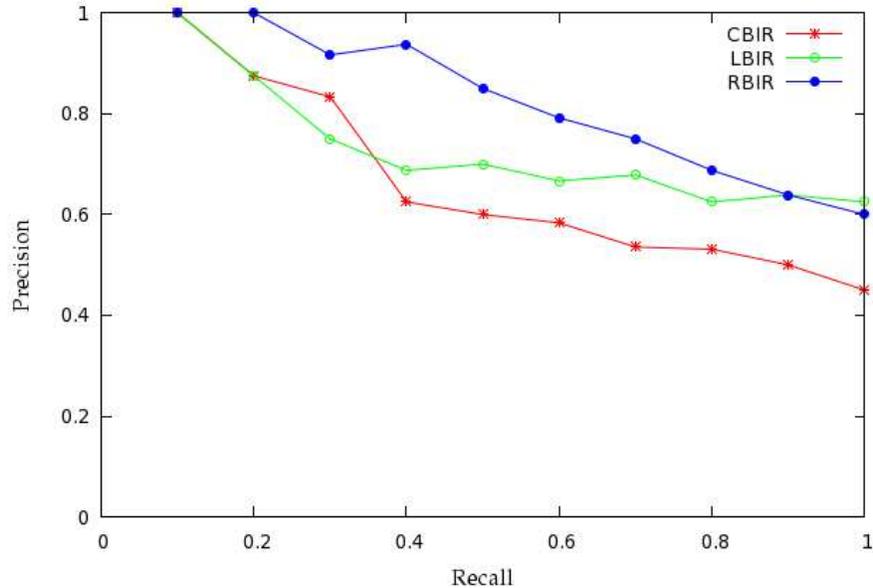}
  \caption{Precision Recall Graph for Top 10 Retrievals}
  \label{fig: Class}
\end{figure*}

\section{CONCLUSIONS}
We have implemented two methods of indexing and retrieval namely: i) region-based indexing and retrieval and ii) location-based indexing and retrieval. 
Hash structure is used to index images. The retrieved images are sorted using Euclidean distance measure in 
the decreasing order of similarity against the query image. The performance of both the systems have 
been measured using standard precision versus recall graphs. Region-based indexing and retrieval gives significantly better results of
81.7 percent precision as compared to location-based indexing and retrieval which gives 72.74 percent
precision. This is because, in most of the cases the tumor is located in the central position. The results are also compared with CBIR which gives 65.33 percent precision.

\small
\balance

\noindent{\includegraphics[width=1in,height=1.7in,clip,keepaspectratio]{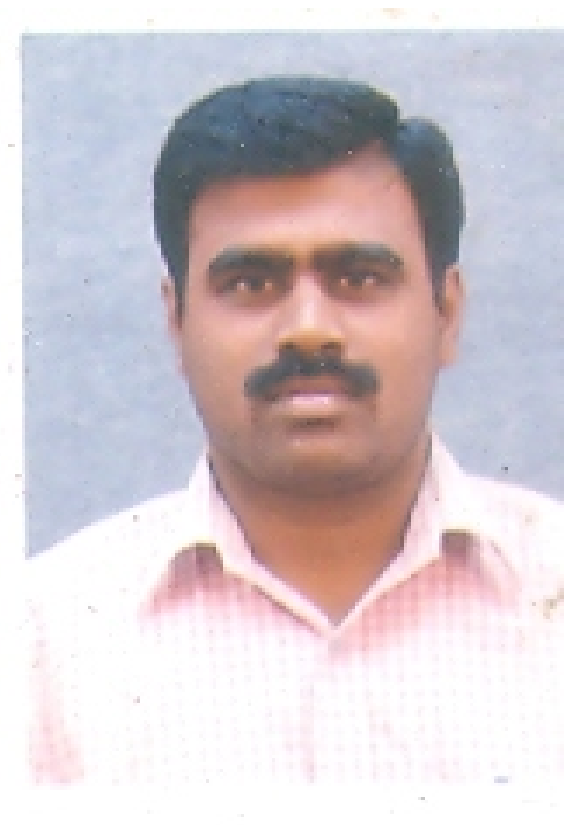}}
\begin{minipage}[b][1in][c]{1.8in}
{\centering{\bf {Krishna A N}} is currently Associate professor, Department of Computer Science and Engineering, S J B Institute of Technology, Bangalore. He obtained his Bachelors and Masters degree in Computer Science and Engineering from University Visvesvaraya College of Engineering. He has publicat-}\\\\\\
\end{minipage}
 ions in International Conferences and Journals. His research interests includes Image Processing, Pattern Recognition and Content Based Image Retrieval. \\\\
\noindent{\includegraphics[width=1in,height=1.7in,clip,keepaspectratio]{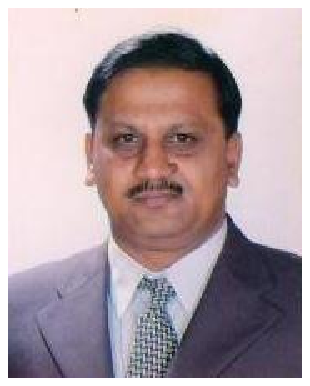}}
\begin{minipage}[b][1in][c]{1.8in}
{\centering{\bf{Dr. B G Prasad}} is currently Professor and Head, Department of Computer Science and Engineering, B N M Institute of Technology, Bangalore. He obtained his Bachelors degree in Computer Science and Engineering from P E S College of Engineering, Mandya, Masters}\\\\
\end{minipage}
degree and Ph.D in Computer Science and Engineering from IIT Delhi. He has publications in many Refereed Journals and Conferences. His research interests include Operating Systems, Content Based Image Retrieval and Computer Networks.

\balance
\end{document}